\documentclass{article}
\usepackage{url}
\usepackage[left=1in, right=1in, top=1in, bottom=1in]{geometry}
\usepackage[utf8]{inputenc} 
\usepackage{graphicx}

\usepackage{draftwatermark}
\SetWatermarkText{PRE-PRINT DRAFT}
\SetWatermarkScale{2}
\SetWatermarkColor[gray]{0.7}

\title{From Ghazals to Sonnets: Decoding the Polysemous Expressions of Love Across Languages}
\author{Sualeh Ali}
\date{\today}

\begin{document}

\maketitle

\begin{abstract}
This paper delves into the intricate world of Urdu poetry, exploring its thematic depths through a lens of polysemy. By focusing on the nuanced differences between three seemingly synonymous words – \textbf{pyaar}, \textbf{muhabbat}, and \textbf{ishq} – we expose a spectrum of emotions and experiences unique to the Urdu language. This study employs a polysemic case study approach, meticulously examining how these words are interwoven within the rich tapestry of Urdu poetry. By analyzing their usage and context, we uncover a hidden layer of meaning, revealing subtle distinctions which lack direct equivalents in English literature. Furthermore, we embark on a comparative analysis, generating word embeddings for both Urdu and English terms related to love. This enables us to quantify and visualize the semantic space occupied by these words, providing valuable insights into the cultural and linguistic nuances of expressing love. Through this multifaceted approach, our study sheds light on the captivating complexities of Urdu poetry, offering a deeper understanding and appreciation for its unique portrayal of love and its myriad expressions.
\end{abstract}

\twocolumn

\section{Literature Review}

There has been a lot of work done in obtaining word embeddings for word sense disambugation. Simplistic approaches exist\cite{sun2017simple} but they require the provision of a \emph{WordNet} which unfortunately for a low resource language like Urdu doesn't exist. There have been attempts however, to come up with a consolidated WordNet like the one put by these authors in Pakistan\cite{zafar2012developing} but these embeddings are either very limited in vocabulary size or are limited in richness of vocabulary for example, only trained on Urdu News Articles. In terms of our main approach which is deducing polysemous words that might occur in the same context we can utilize many approaches such as dependency sequencing\cite{chen2009fully}, but we will be using a modification of the famous Latent Dirichlet allocation approach to impute word senses by Blei\cite{blei2003latent}.

\section{Introduction}
Urdu is a language rich in polysemous meanings of text. This case study will follow the usages of three sense collocations of love expressed in this study as (\textbf{LOVE\textsubscript{1}} vs \textbf{LOVE\textsubscript{2}} vs \textbf{LOVE\textsubscript{3}}). These will represent in the same order:

\textbf{Pyaar} - Taken from Sanskrit word (priya) in Hindi. It is used to describe almost all kinds of love whether it is your lover, parent, children, friends, or anyone. It is a less formal word; its formal word from the same origin can be \textbf{Sneh} (to describe lustless love towards any people) or \textbf{Prem} (love for your lover, it may or may not be lustless). 

\textbf{Ishq} - Taken from Persian word (ishq) Urdu. It is used to describe extreme love (lustless love) towards someone, which can be your lover, Prophet, or God. 

\textbf{Mohabbat} - Taken from Arabic word (hubb) Urdu. It is used to describe your love towards your children, lover (may or may not be lustless), or spiritual. It is a more formal word than the others.

\section{Method}

In this section, we detail the methodology employed to analyze word embeddings of Urdu and English words, focusing on the semantic space, similar word senses, and their distribution.

\section{Word Embeddings}

\subsection{Dataset Curation}
Unfortunately, due to the unavailability of a \emph{WordNet} for Urdu as is the case working with low-resource languages we had to curate our own dataset. We utilized a corpus provided by the Rekhta organization, focusing on Urdu poetry \cite{Rekhta}. To facilitate a fair comparison, we employed a similar embedding approach for English using the Shakespeare sonnet data-set curated by Edmond\cite{edmondson2004shakespeare}. Before using this, we had to filter out old English stop words such as \emph{"thou,"} \emph{"thee,"} and others were considered as stop words to account for language evolution.

\subsubsection{Word Sense of LOVE}
We identified three similar word senses of love for English after examining the synsets in WordNet: love, passion, and affection. These senses were chosen due to their significant overlap and potential differences in usage.

\begin{figure}[h]
    \centering
    \includegraphics[width=0.8\linewidth]{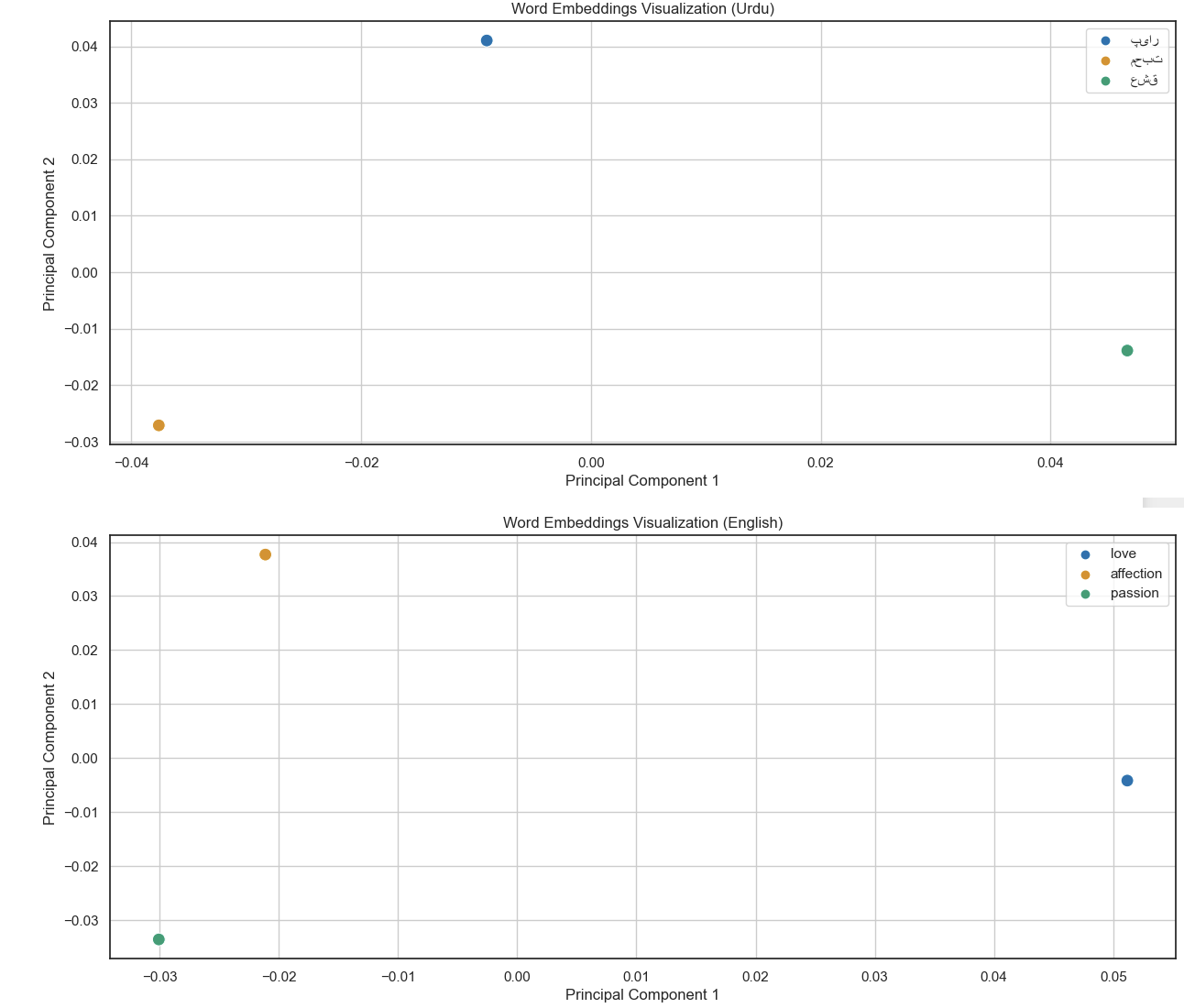}
    \caption{Mapping of Urdu and English word vectors in the semantic space.}
    \label{fig:semantic_space}
\end{figure}

\subsection{Visualization of Word Vectors}
Principal Component Analysis (PCA) was used to project the generated word vectors into a lower-dimensional semantic space. This facilitated visualization and analysis of the relationships between different word senses. Cosine similarity scores were calculated to assess the semantic similarity between words within each word sense and across languages. This provided insights into the usage patterns of each sense in English and Urdu.

\begin{figure}[h]
    \centering
    \includegraphics[width=0.8\linewidth]{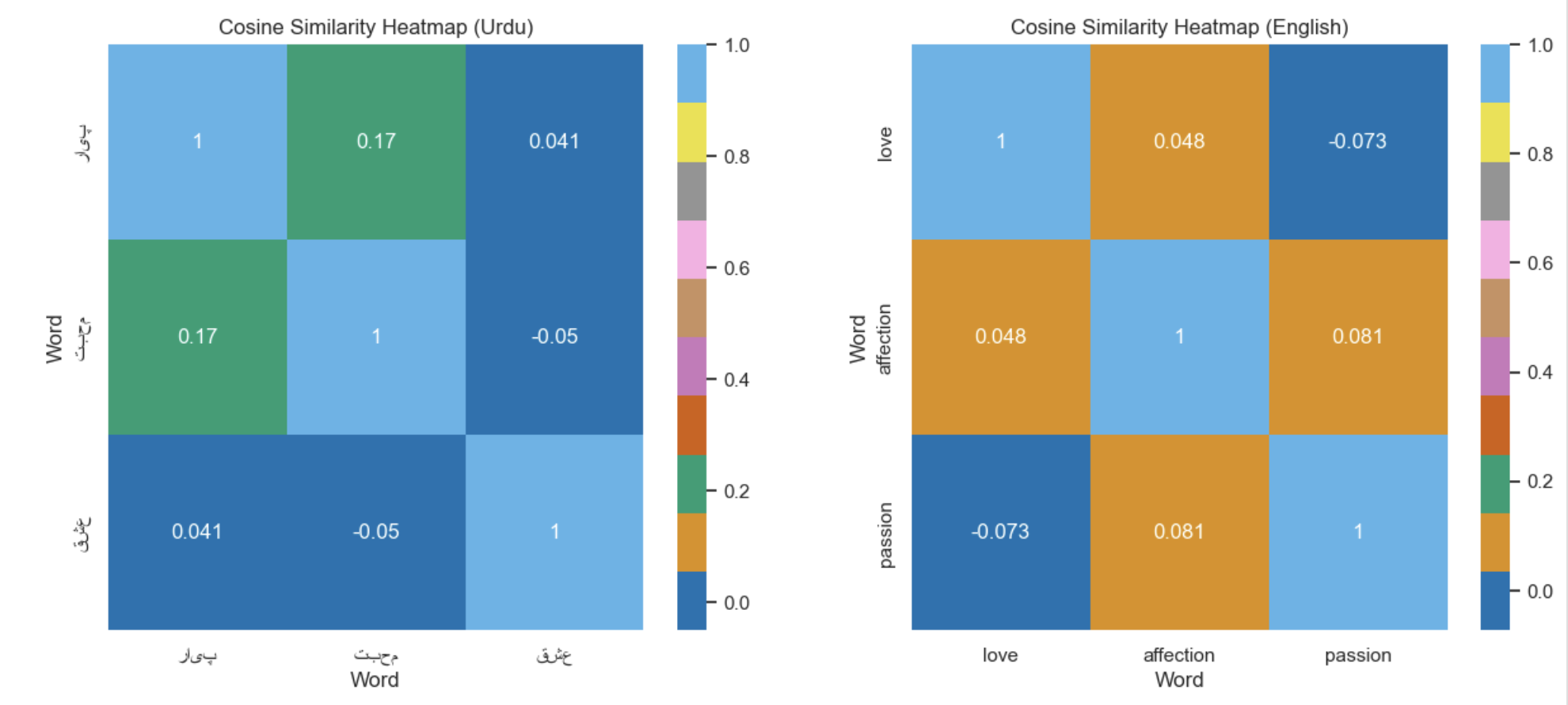}
    \caption{Comparison of cosine similarity scores between Urdu and English word senses.}
    \label{fig:cosine_similarity}
\end{figure}

The interesting thing to note here is that some words over here in English are negatively correlated suggesting the opposite in some case. However, both English and Urdu do not share strong similarity scores suggesting that their usages are quite different.

\subsubsection{Synsets of LOVE Words}
We further analyzed the similar words for each word sense (LOVE\textsubscript{1}, LOVE\textsubscript{2}, LOVE\textsubscript{3}) in both Urdu and English.

\begin{figure}[h]
    \centering
    \includegraphics[width=0.8\linewidth]{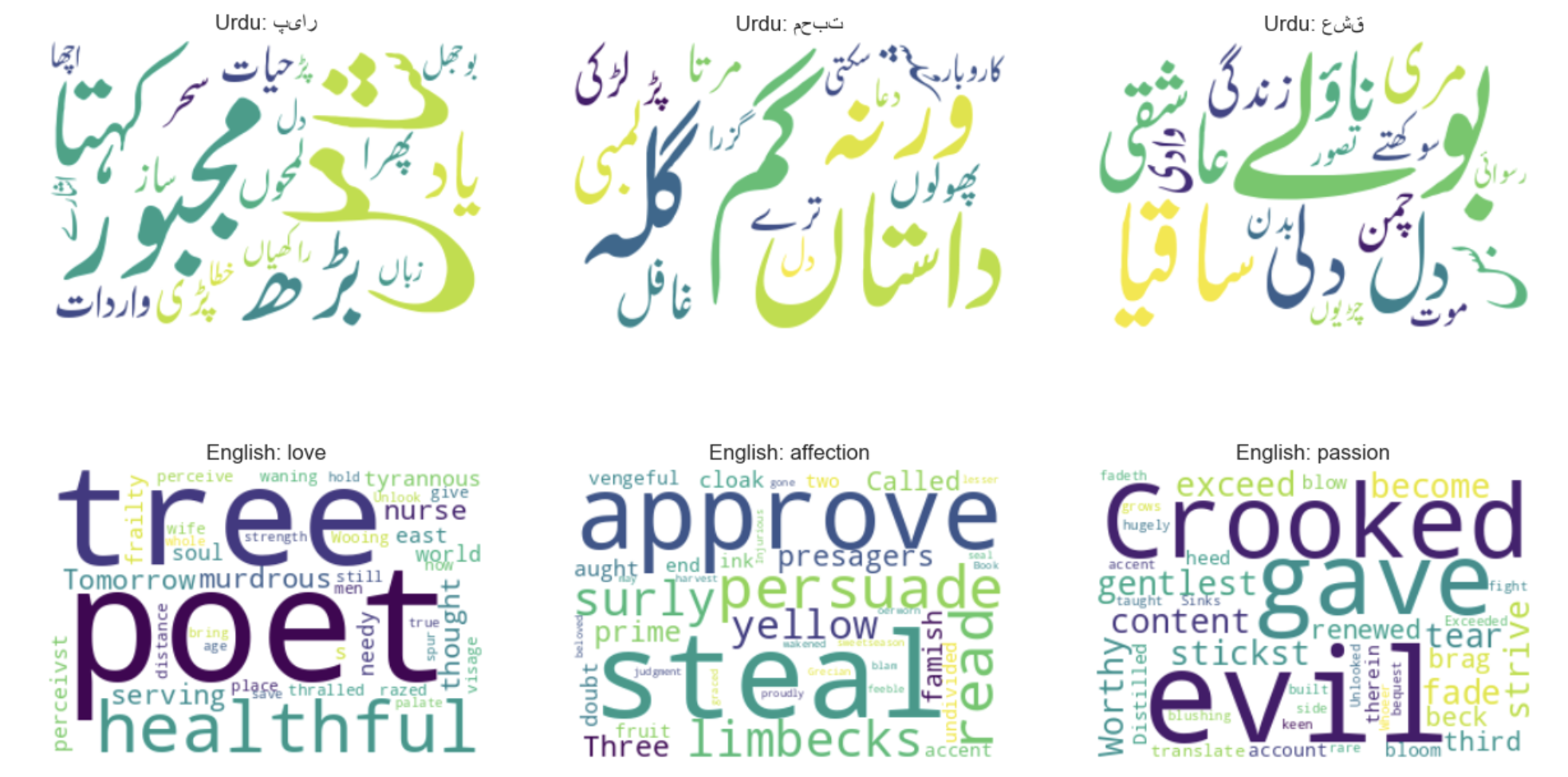}
    \caption{Word clouds depicting the synsets of LOVE words in Urdu and English.}
    \label{fig:word_clouds}
\end{figure}

\begin{figure}[h]
    \centering
    \includegraphics[width=0.8\linewidth]{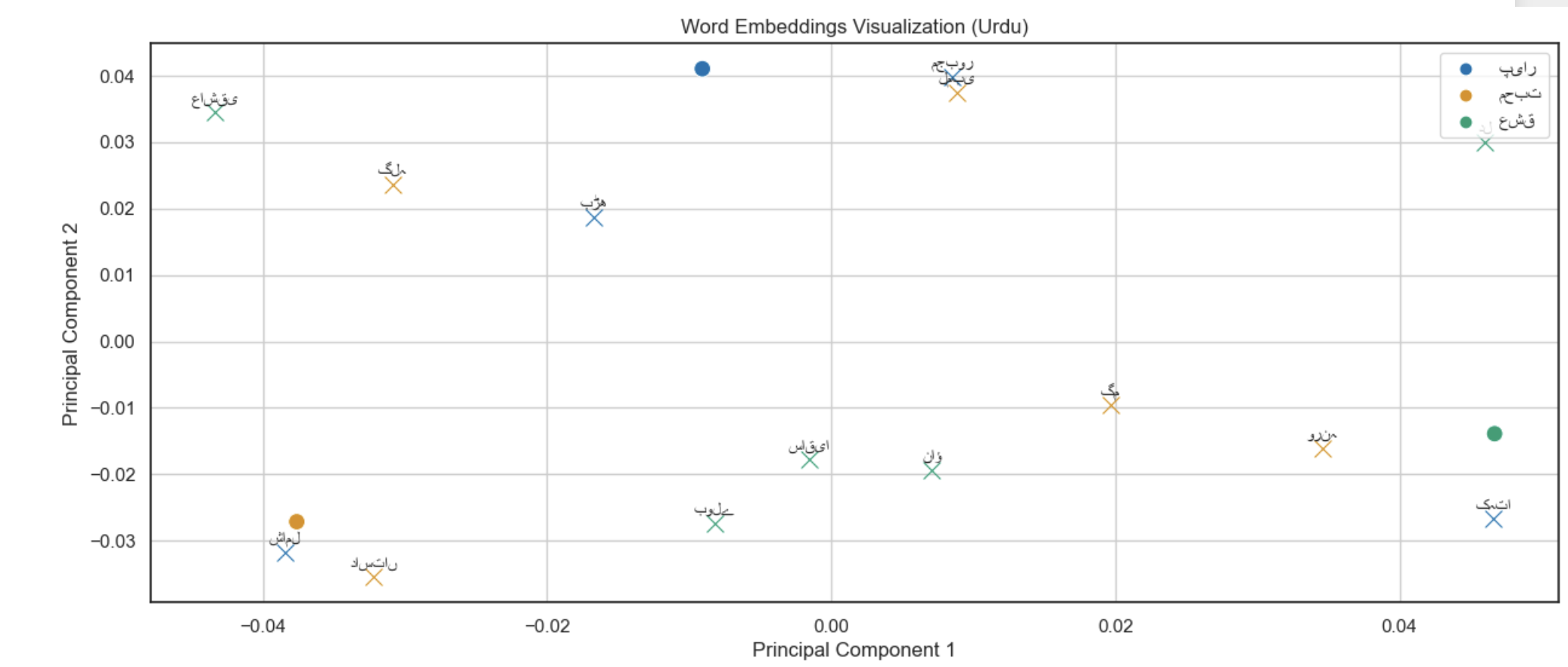}
    \caption{Mapping of Urdu similar word embeddings.}
    \label{fig:urdu_similar_mapping}
\end{figure}

\begin{figure}[h]
    \centering
    \includegraphics[width=0.8\linewidth]{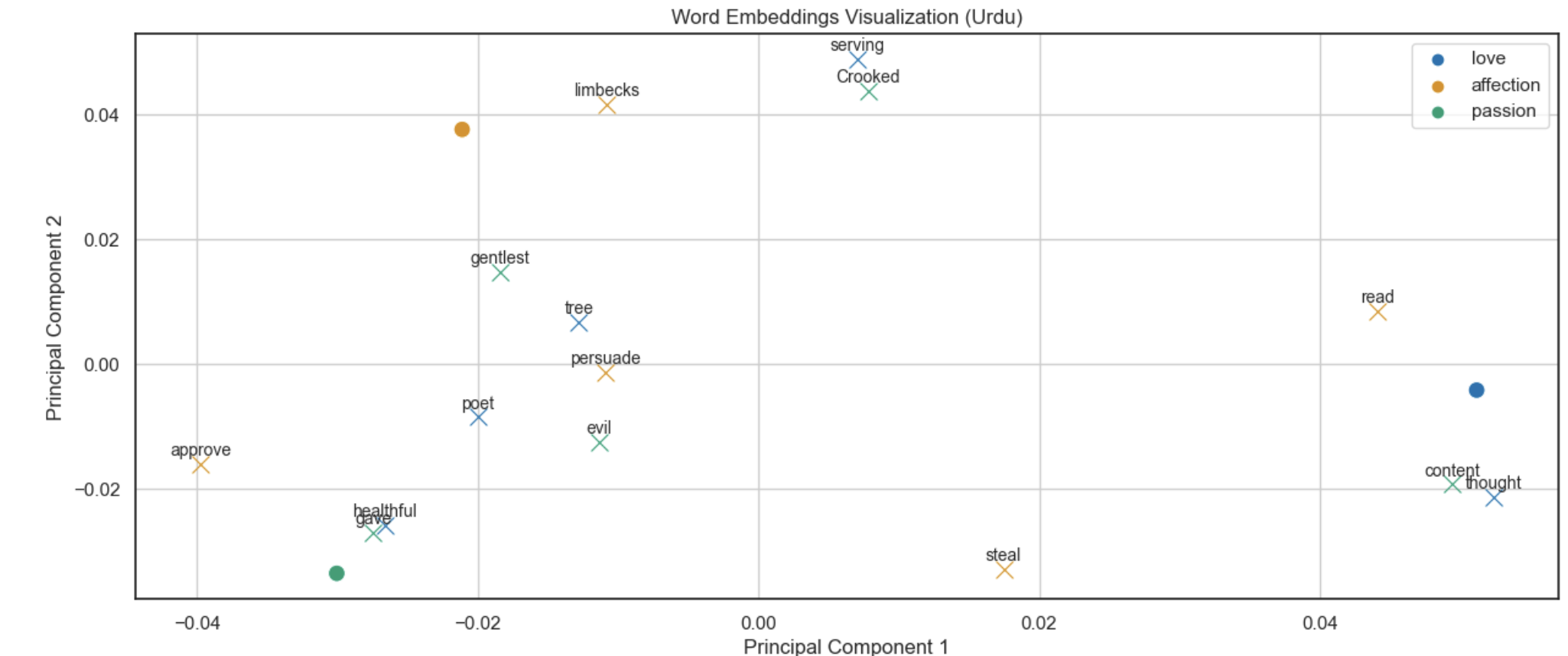}
    \caption{Mapping of English similar word embeddings.}
    \label{fig:english_similar_mapping}
\end{figure}

\section{Mapping Word Embeddings}
To understand the overlap and distribution of similar words in the embedding space, we mapped all word embeddings in the same space.

\subsubsection{Observations}
\begin{itemize}
    \item \textbf{Urdu Word Vectors:} Exhibit scattered distribution with uncertain cluster membership, suggesting overlap with other polysemous variations. The green cluster, representing LOVE\textsubscript{2} (Ishq), stands out as the closest and strongest group.
    \item \textbf{English Word Vectors:} Demonstrate significant overlap between senses, particularly in the green and blue clusters, corresponding to LOVE\textsubscript{2} and LOVE\textsubscript{3} (passion and affection), indicating shared membership.
\end{itemize}

\section{Word Sense Induction}
Formally the problem is known as Word Sense Disambiguation, pioneered by Yarowsky\cite{yarowsky1995unsupervised}, but since we are working with low-resource languages we call this \textit{Word Sense Induction}. There is plenty of research that has gone into this, such as context clustering, word clustering, and co-occurrence graph. We will be opting for a relatively simpler solution.

Considering the fact that there is a considerable overlap between some senses of the Urdu Words, we would like to employ some unsupervised algorithm that tries to disambguate distinct word senses and find out that if within these collocations are they separate uses of love that might exist. For this the first thing we need to do is mask out all of the collocations of the word love, however since we need to underpin what the underlying collocation might be, we preserve its original sense information with the target token *LOVE.n.1* For example, the Romanized version of the provided Urdu text, with the target token \emph{LOVE.n.1} included, is:

\begin{quote}
Dil [MASK] ki nazar ke liye beqarar hai
Ek teer is taraf bhi, yeh taza shikaar hai
Muhabbat ki har ek adaa, ik adhbut kahaani hai
Jaise kehni koi aap se seekhe, yeh taza shikaar hai
\end{quote}

Once we have masked out the token, we apply the very well reputed technique of \emph{Latent Dirichlet allocation (LDA)}; which assumes the data comes through a probabilistic process, which might very much be the case when a poet is trying to utilise one of the word senses in their toolbox. But before we apply LDA we also need to filter all of the corpus that contain \emph{any} of the word senses. Also note however, if a particular couplet contains more than one word sense then we will assume the first one collocation which is an assumption in this case. The following results show that there at best, six topics that might capture the usage of different collocations in six senses. Also given the fact that we get a coherence score of 0.59 rounded to two significant figures, which is not bad considering the subjective nature of the dataset. We also then try to further understand how each of these original word senses might be distributed across these topics.

\begin{figure}[h]
    \centering
    \includegraphics[width=0.8\linewidth]{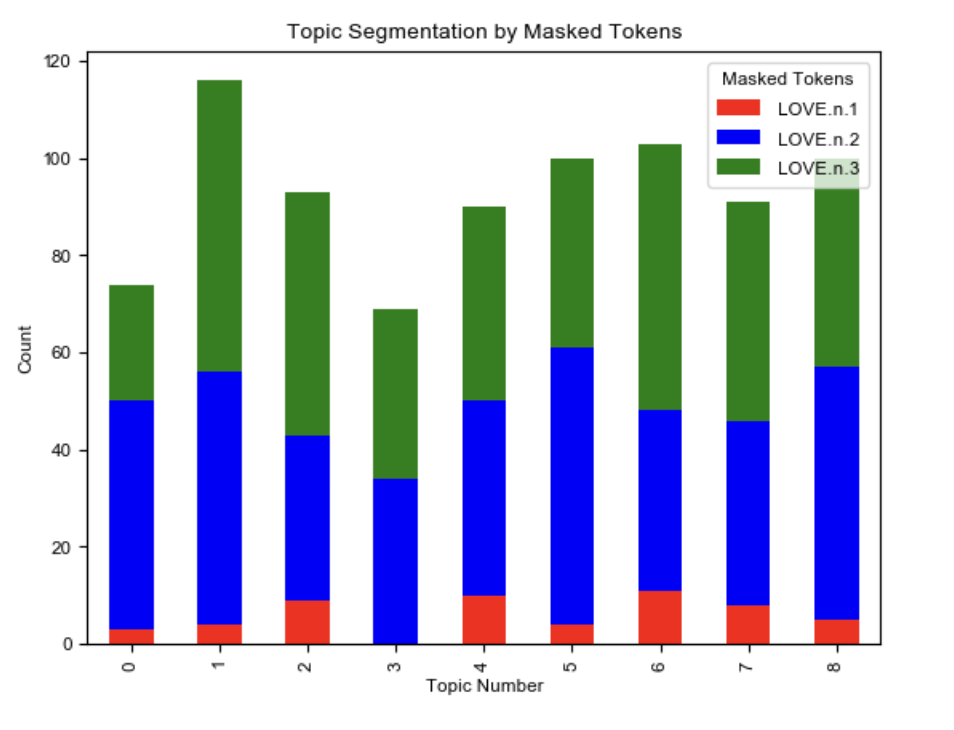}
    \caption{Topic Segmentation of LOVE senses in Urdu}
    \label{fig:topic_segmentation_plot}
\end{figure}

We can see the usage of all three sense is relatively the same however there are some topics with interesting observations for example \emph{Topic 3} has a topic distribution in which \emph{ishq} usage is not there. Also to extrapolate this we see that LOVE\textsubscript{1} and LOVE\textsubscript{2} dominate most of the topic with a very sparse distribution of Topic 3.

\section{Results}

\subsection{Love}
\begin{tabular}{|c|c|}
    \hline
    \textbf{Word} & \textbf{Score} \\
    \hline
    poet & 0.35 \\
    tree & 0.33 \\
    healthful & 0.32 \\
    serving & 0.31 \\
    thought & 0.31 \\
    \hline
\end{tabular}

\subsection{Affection}
\begin{tabular}{|c|c|}
    \hline
    \textbf{Word} & \textbf{Score} \\
    \hline
    steal & 0.31 \\
    approve & 0.31 \\
    persuade & 0.31 \\
    read & 0.30 \\
    limbecks & 0.30 \\
    \hline
\end{tabular}

\subsection{Passion}
\begin{tabular}{|c|c|}
    \hline
    \textbf{Word} & \textbf{Score} \\
    \hline
    evil & 0.33 \\
    Crooked & 0.30 \\
    gave & 0.29 \\
    gentlest & 0.29 \\
    content & 0.28 \\
    \hline
\end{tabular}

\subsection{Pyar}
\begin{tabular}{|c|c|}
    \hline
    \textbf{Word} & \textbf{Score} \\
    \hline
    majboor & 0.41 \\
    shamil & 0.40 \\
    kehta & 0.39 \\
    barh & 0.36 \\
    yaad & 0.35 \\
    \hline
\end{tabular}

\subsection{Muhabbat}
\begin{tabular}{|c|c|}
    \hline
    \textbf{Word} & \textbf{Score} \\
    \hline
    gum & 0.41 \\
    dastaan & 0.38 \\
    warna & 0.37 \\
    gala & 0.36 \\
    lambi & 0.36 \\
    \hline
\end{tabular}

\subsection{Ishq}
\begin{tabular}{|c|c|}
    \hline
    \textbf{Word} & \textbf{Score} \\
    \hline
    bole & 0.46 \\
    saqia & 0.39 \\
    ishq & 0.37 \\
    nao & 0.36 \\
    dil & 0.36 \\
    \hline
\end{tabular}

\vspace{\baselineskip}

Each table shows the top five words similar to the sense (LOVE\textsubscript{1}, LOVE\textsubscript{2}, LOVE\textsubscript{3}) with their respective scores.

\subsubsection{Urdu Analysis}
The analysis of LOVE words in Urdu reveals distinct connotations associated with each word sense:

\begin{itemize}
    \item \textbf{LOVE\textsubscript{1} (Pyar):} Expresses ideas related to personal relationships, encompassing words like \textit{yaad} (memory) and \textit{dil} (heart).
    \item \textbf{LOVE\textsubscript{2} (Ishq):} Portrays deeper emotions with words such as \textit{aashiqui} (beloved or devotee), \textit{zindagi} (life), and \textit{mout} (death).
    \item \textbf{LOVE\textsubscript{3} (Muhabbat):} Carries a melancholic undertone with frequent mentions of romantic interests like \textit{larki} (girl) and \textit{pholoon} (flowers).
\end{itemize}

\subsubsection{English Analysis}
Contrastingly, the connotations of LOVE words in English exhibit unpredictability:

\begin{itemize}
    \item \textbf{LOVE\textsubscript{1} (Love):} Encompasses diverse themes such as \textit{poet}, \textit{tree}, and \textit{tomorrow}, reflecting a wide spectrum of associations.
    \item \textbf{LOVE\textsubscript{2} (Affection):} Introduces words like \textit{approve}, \textit{persuade}, and connotations of vengeance, suggesting a blend of romantic and thematic elements.
    \item \textbf{LOVE\textsubscript{3} (Passion):} Features quirky words like \textit{crooked}, \textit{evil}, and \textit{gentlest}, hinting at potential negative connotations within the word's semantic space.
\end{itemize}

\begin{table}[h]
    \centering
    \begin{tabular}{|c|c|}
        \hline
        \textbf{Number of Topics} & \textbf{Coherence Value} \\
        \hline
        2 & 0.485745 \\
        3 & 0.545730 \\
        4 & 0.524656 \\
        5 & 0.574208 \\
        6 & \textbf{0.589357} \\
        7 & 0.584264 \\
        8 & 0.570953 \\
        9 & 0.574424 \\
        10 & 0.584227 \\
        \hline
    \end{tabular}
    \caption{Coherence values for different numbers of topics in the LDA model. The highest coherence value is achieved when the number of topics is 6.}
    \label{tab:lda_coherence}
\end{table}

\section{Discussion}
Urdu word vectors exhibited greater sparsity and less clear clustering patterns compared to English. This suggests a potentially wider semantic range and less consistent usage of the word "love" in Urdu. English word vectors demonstrated a more unpredictable distribution across different senses. This indicates a potentially more nuanced and complex usage of the word "love" in English. The analysis highlights the importance of considering the cultural and linguistic context when interpreting the semantic space of emotions like "love."

\section{Limitations}
Our approach, while yielding valuable insights, is not without its limitations. These considerations should be taken into account to contextualize the scope and applicability of our findings.
\textbf{Fixed Number of Topics/Senses:} The method presupposes a predefined number of topics or senses, assuming that this choice accurately encapsulates the intricate semantic variations present in the data. However, the fixed nature of this parameter might oversimplify the dynamic and multifaceted nature of word meanings.
\textbf{Single Sense Collocation Per Row:} The methodology operates under the assumption that each row in the dataset corresponds to a distinct sense or meaning of a word. This simplification may not adequately capture the polysemy and nuanced variations in word usage, potentially leading to misrepresentations of the underlying semantic space.
\textbf{Lack of Metadata:} The absence of additional metadata, such as part-of-speech (POS) information, limits the depth of our analysis. Integrating metadata could enrich our understanding of semantic nuances associated with different word senses, providing essential context to our interpretations.
\textbf{Limited Explainability:} Interpretability of our findings is constrained by the inherent subjectivity in defining semantic senses. While our methodology generates quantitative results, the underlying semantic relationships may not always align seamlessly with human intuition, highlighting the challenge of achieving explain-ability in semantic sense analysis. While our methodology provides valuable insights, researchers should consider these limitations when interpreting and extrapolating the results to different linguistic and contextual scenarios.

\section{Conclusion}
The proposed methodology for sense induction and similarity analysis proves effective, particularly in low-resource language scenarios without WordNet availability. The coherence analysis reveals optimal performance with six topics, highlighting the significance of parameter tuning. Exploring LOVE\textsubscript{1}, LOVE\textsubscript{2}, and LOVE\textsubscript{3} across Urdu and English unveils nuanced semantic variations. The visualizations, word clouds, and scatter plots offer comprehensive insights into word embeddings and contextual relationships. While successful, certain limitations exist, such as the need for a predefined number of topics and reliance on single-sense collocation. Future research could explore enhancements and incorporate additional indicators for model refinement. In summary, a proposed approach serves as a versatile tool for semantic analysis, especially in languages with limited lexical resources. The combination of coherence analysis and detailed sense exploration provides valuable insights, paving the way for future advancements in the field.

\section*{Acknowledgement}

We would like to thank the Rekhta Organization\cite{Rekhta} for the facilitation of the data set without which the research wouldnt have been possible and also a curation of Urdu stop words we found online\cite{urdu-stopwords}.

\end{document}